\documentclass[10pt]{article} 
\usepackage[preprint]{tmlr}

\usepackage{amsmath,amsfonts,bm}

\def\eqref#1{equation~\ref{#1}}

\def\1{\bm{1}}

\DeclareMathAlphabet{\mathsfit}{\encodingdefault}{\sfdefault}{m}{sl}
\SetMathAlphabet{\mathsfit}{bold}{\encodingdefault}{\sfdefault}{bx}{n}

\usepackage{amsthm}
\newtheorem{lemma}{Lemma}
\usepackage{graphicx}
\usepackage{booktabs}
\usepackage{hyperref}
\usepackage{url}

\title{Gradient Prediction with Control Variates \\ in the Cheap-Forward Regime}

\author{\name Kamil Ciosek 
      \email kamilc@spotify.com\\
      \addr Spotify
      \AND
      \name Nicol\`o Felicioni
      \email nicolof@spotify.com  \\
      \addr Spotify
      \AND
      \name Juan Elenter
      \email juane@spotify.com \\
      \addr Spotify
      \AND
      \name Ehsan Imani
      \email ehsani@spotify.com \\
      \addr Spotify
      }

\def\month{MM}  
\def\year{YYYY} 
\def\openreview{\url{https://openreview.net/forum?id=XXXX}} 

\begin{document}

\maketitle

\begin{abstract}
We study whether otherwise-idle inference resources could reduce the scarce-GPU cost of training. Our analysis uses a simulated compute ledger in which fleet work is billed at a fraction of a scarce-GPU forward; all experiments run on a regular GPU. Our algorithm predicts gradients with a reduced-precision, inference-style reverse-mode program and combines many predictions with a few exact gradients through a control variate, so approximation error becomes variance rather than bias. On a 124M-parameter language model and selected short training windows, the method can lower simulated ledger cost relative to the tested baselines when fleet work is sufficiently cheap. Experiments spanning 10M--774M parameters show both transfers and failures. We do not test inference-only hardware, end-to-end distributed latency, or a full optimizer-by-batch-size baseline sweep.
\end{abstract}

\section{Introduction} \label{sec:intro}

We study a hypothetical setting in which otherwise-idle inference resources are available while training relies on backward-capable GPUs. We ask whether such resources could reduce training compute cost under a simulated ledger with billing rate $\gamma$, where fleet work is charged at a fraction $\gamma$ of a scarce-GPU forward. We use $\gamma \to 0$ as an idealized free-fleet limit, $\gamma = 0.01$--$0.1$ as illustrative discounted rates, and $\gamma = 1$ as an equal-hardware re-billing control. All experiments execute both trainer and fleet on a regular GPU; no result measures actual inference-only hardware or end-to-end distributed time.

Two ideas turn cheap forwards into training progress. First, the fleet \emph{predicts} gradients. We hand-roll a backward pass out of nothing but reduced-precision, forward-style operations: matrix multiplies, elementwise maps, and recomputation; no autograd and no training-sized memory. This gives the predictor an inference-like operation set and memory profile, but does not establish that current inference-only hardware or serving stacks can execute it. The predictions are good in that they explain about 98\% of per-example gradient variance on our 124M-parameter testbed. However, they are cheap for a reason: int8 arithmetic and half-precision streams all inject systematic error, and at larger scales activation outliers can break low-precision matmuls outright \citep{inteight}. No amount of averaging removes that kind of error. Second, the trainer repairs the bias with a control variate \citep{kleijnen1975statistical}. Each step it computes exact gradients for a small control batch, has the fleet predict gradients for both the controls and a much larger prediction batch, and adds the \emph{difference} of the two predicted means to the exact mean; this is the same difference trick that powers variance-reduced optimization \citep{NIPS2013_ac1dd209}. Whatever systematic error the predictor has appears in both predicted means and cancels, so the gradient estimate stays unbiased no matter how rough the prediction is: approximation error becomes variance rather than bias. Variance is a currency we know how to price as it trades against batch size \citep{large-batch}; how much of the reduction becomes actual progress depends on the optimizer consuming it; we demonstrate that using Muon \citep{jordan2024muon} converts nearly all of it.

Both ingredients differ from existing work. Zeroth-order fine-tuning also trains with forwards alone \citep{mezo}, but its gradient guesses carry variance that grows with model size, whereas our predictions are high-fidelity gradients computed a cheaper way. Low-communication distributed training spreads work across many machines \citep{diloco}, but every worker still runs a true backward
on training-capable hardware; our fleet never does.

\textbf{Contributions.} (i)~A cost model for the cheap-forward regime and a calculus that converts predicted-gradient quality into an equivalent batch size and a simulated ledger cost. (ii)~A gradient predictor designed around reduced-precision, inference-style operations and memory use, evaluated on a GPU. (iii)~An estimator that combines predicted and exact gradients with a control variate, provably unbiased for \emph{any} predictor under the stated sampling and coefficient assumptions, with an online coefficient that self-tunes and can attenuate a failing predictor. (iv)~GPU measurements showing lower simulated ledger cost than the tested, non-exhaustive baselines in some short-window regimes, together with a map from 10M to 774M parameters of where the approach transfers and where it fails.

\section{Preliminaries}
\label{sec:prelim}

\paragraph{Control variates.} At a training step we want the population gradient $\mu = \mathbb{E}[g(x)]$, where $g(x)$ is the exact per-example gradient. The plain estimate is the mean $\bar g_C$ over a batch $C$ of $m_c$ examples; its error is pure sampling noise.  Now suppose we also have a cheap \emph{approximate} gradient $h(x)$ that tracks $g(x)$ example by example, and that we can afford to evaluate it on a much larger batch $P$ of $m_p$ examples. The control-variate estimator
\begin{equation}
\hat\mu_\beta \;=\; \bar g_C \;+\; \underbrace{\beta\,\bigl(\bar h_P - \bar h_C\bigr)}_{\text{correction}}
\label{eq:cv}
\end{equation}
adds a correction with two properties. First, its mean is zero. This is because $C$ and $P$ are drawn from the same distribution, so any systematic error of $h$ appears in both terms and cancels. This leaves $\hat\mu_\beta$ unbiased for \emph{any} $h$ and any fixed $\beta$. And it is anti-correlated with the noise in $\bar g_C$: because $h$ mimics $g$, the term $-\beta \bar h_C$ cancels part of the small batch's sampling fluctuation, while $+\beta \bar h_P$ restores the same signal measured on many more examples, where it is far less noisy. A useful way to
read the mixing coefficient: if the approximation were \emph{perfect} ($h = g$), choosing $\beta = m_p/(m_p + m_c)$ makes Equation~\ref{eq:cv} collapse to the plain mean over the concatenation $C \cup P$---training on both batches as one. In other
words, $\beta$ is the share of the total sample that the cheap batch deserves: with a perfect predictor its examples count fully, and as prediction quality degrades the best $\beta$ shrinks, discounting them toward zero. How to set $\beta$ from measured quantities is part of our method (Section~\ref{sec:method}).

\paragraph{Cost model.} On a training GPU the backward pass costs about twice a forward, so the standard rule of thumb prices one training step at roughly three forward-equivalents; we adopt the forward-equivalent as the unit of compute throughout, measuring every operation's cost as wall-clock relative to one forward on the scarce GPU at the same batch size. The cheap-forward regime adds one parameter to this model: a forward executed on the modeled fleet is billed at a fraction $\gamma \le 1$ of a scarce-GPU forward. The ledger (simulated) cost of a run is then its scarce-GPU time plus $\gamma$ times the fleet's metered forward-equivalents, and a method helps only if it reaches the target loss at lower ledger cost than exact training. All of our claims are settled at explicit values of $\gamma$; since
each run logs its own meters, any run can be re-billed at any $\gamma$ after the fact.

\section{Method} 
\label{sec:method}
\subsection{Cheap Forward: predicting gradients with inference-style operations}\label{sec:cheap-forward}
We target an inference-style op set consisting of matrix multiplies, elementwise maps, and reductions, without autograd or a training-sized activation stack. Our GPU implementation hand-implements the backward around that op set, subject to the fused-attention caveat below. Nothing about the formulas is mathematically approximate; however, executing them in reduced precision, at weights a few steps stale, makes the result an approximation $h(x) \approx g(x)$; we call it a prediction and lean on the estimator of Section~\ref{sec:cv} for correctness.

\paragraph{The transposed Jacobian.}
Backpropagation is nothing more than applying each layer's \emph{transposed Jacobian} to an error signal, layer by layer from the loss back to the parameters. The reason this may be feasible on an inference-style executor is that, for every layer in a transformer, the transposed Jacobian is a closed-form program built from the same primitives as the layer's forward. For a linear layer it is a matrix multiply with the transposed weight; for attention it is a handful of matrix multiplies against cached queries, keys, values, and attention probabilities, plus the softmax derivative as an elementwise map; for GELU it is one elementwise multiply. The pass starts at the head, where the cross-entropy gradient is available in closed form (softmax minus one-hot), and pulls this signal backward through each block; a parameter's per-example gradient is then the outer product of two streams the pass already has: the layer's cached input and the arriving error signal. 

\paragraph{What makes predictions cheap.} A direct transcription of the above would be correct but would neither fit in inference memory nor beat a GPU on cost. The predictor adds five optimizations (we do not claim these optimizations are original to our work except in the specific context of control variates):
\begin{itemize}
\item \emph{Recompute instead of store.} A backward pass needs the intermediate values that the forward computed inside every layer,
  and storing them all is exactly the memory bill that separates training from inference. Instead, we cut the network into $R$
  segments and keep only its running state at the segment boundaries ($R{+}1$ small tensors). The backward then visits the segments in
  reverse order: it re-runs one segment's forward from its saved boundary to regenerate the values inside, uses them, and discards
  them before moving on. Memory stays near inference level i.e the boundaries plus a single segment's internals. The computational price is one
  extra forward pass. 
\item \emph{Streamed head.} The vocabulary-sized tensors at the loss layer are produced in chunks over the batch, so the full
  logits matrix never exists at once.
\item \emph{Factored per-example gradients.} For a weight matrix, one example's gradient is the product of two thin matrices the pass
  already has, specifically the layer's cached input $X_i$ and the arriving error signal $D_i$, one row per token: $dW_i = X_i^\top D_i$. We keep
  these factors and never form the $dW_i$ themselves. Not because the optimizer could consume a factored form (it takes one dense
  matrix), but because nothing downstream needs \emph{per-example} gradients as matrices. Only two things are ever read off: the batch
  mean $\tfrac1B \sum_i X_i^\top D_i$, a single dense product over the stacked factors, and the per-example covariance statistics of
  Section~\ref{sec:cv}, which involve only inner products between gradients; an inner product of two outer products collapses to
  small token-Gram matrices,
  \begin{equation}
  \langle dW_i,\, dW_j\rangle
  \;=\; \sum_{t,s}\,
  (x_{i,t}^{\!\top} x_{j,s})\,(d_{i,t}^{\!\top} d_{j,s}),
  \label{eq:gram}
  \end{equation}
  a $T \times T$ computation instead of a weight-sized one. Skipping the $B$ dense per-example matrices, each as large as the weight itself, is what keeps per-example capture inside inference memory. On the large prediction batch even the factors are dropped: only
  the batch-sum gradient is accumulated, one GEMM per weight.
\item \emph{Fused attention kernels, called directly.} Our measured GPU configuration invokes fused attention forward and backward kernels as plain functions (i.e.\ without an autograd graph), reducing the predictor's cost relative to the materialized-probabilities chain. The forward kernel is commonly available in inference stacks; the backward kernel currently ships only with training frameworks. We include its GPU-measured cost in the simulated ledger because it is a fixed sequence of tiled matrix-multiply and elementwise operations with inference-like memory use, but we have not shown that current inference-only hardware or serving stacks can execute it. Supporting it may require nontrivial kernel and systems work.
\item \emph{int8 matrix multiplies.} Trunk weight GEMMs run on int8 tensor cores with symmetric scales, applied per token for activations and error signals, per output channel for weights, the latter quantized once per weight sync. Attention's activation-activation products, the softmax statistics, and the vocabulary head stay in half or single precision, where int8 buys nothing. On our from-scratch models this recipe preserves fidelity ($h$ explains $\hat\rho^2 \approx 0.98$ of per-example gradient variance at 124M); on large \emph{pretrained} checkpoints, activation outliers break per-token scaling \citep{inteight} and the fleet falls back to bf16 matmuls with a single-precision residual stream.
\end{itemize}

\paragraph{Cost and verification.} We meter the predictor in forward-equivalents: $c_h$ is its wall-clock per example divided by that of one model forward on the same GPU and batch. Counting GEMMs (forward, recompute, error-signal, weight-gradient) bounds $c_h$ from below at about $3$; measured values about $12$ at 410M-774M, so predicting one example's gradient costs $\gamma\,c_h$ scarce-forward units on the ledger. Because the whole pass is deterministic arithmetic, it admits a sharp gate: run in fp64 with quantization off, it must reproduce autograd's per-example gradients to relative error $10^{-12}$ (measured between $\sim\!10^{-15}$ and $10^{-13}$ on all model families). 

\subsection{Optimal Mixing of The Control Variate}
\label{sec:cv}
Section~\ref{sec:prelim} established that the estimator of Equation~\ref{eq:cv} is unbiased for \emph{any} predictor $h$ and any fixed $\beta$. The mixing coefficient $\beta$ therefore has exactly one job: minimize the estimator's noise. Since the mean is always right, we measure noise as the expected squared error $V(\beta) = \mathbb{E}\,\lVert \hat\mu_\beta - \mu \rVert^2$ and ask which $\beta$ makes it smallest.

The answer depends on three scalars that summarize how the per-example vectors $g(x)$ and $h(x)$ fluctuate around their means when $x$ is drawn from the data distribution:
\begin{equation}
\Sigma_g = \mathbb{E}\,\bigl\lVert g - \mathbb{E}g \bigr\rVert^2,
\qquad
\Sigma_h = \mathbb{E}\,\bigl\lVert h - \mathbb{E}h \bigr\rVert^2,
\qquad
C_{gh} = \mathbb{E}\,\bigl\langle g - \mathbb{E}g,\;
h - \mathbb{E}h \bigr\rangle,
\label{eq:moments}
\end{equation}
with $g$ and $h$ evaluated on the \emph{same} example. These are analogous to the usual variance and covariance, but totaled over the coordinates of a vector, and they behave the way the scalar versions do.\footnote{They are bilinear, they add across independent quantities, and Cauchy--Schwarz bounds the correlation $\rho = C_{gh}/\sqrt{\Sigma_g \Sigma_h} \in [-1, 1]$, which measures how faithfully the prediction tracks the exact gradient.}

\begin{lemma}[Optimal mixing]\label{lem:beta}
For the estimator of Equation~\ref{eq:cv}, with $C$ (control micro-batch used for exact gradients) and $P$ (prediction micro-batch used for approximate gradients) drawn
independently from the data distribution,
\begin{equation}
V(\beta) \;=\; \frac{\Sigma_g}{m_c}
\;-\; 2\beta\,\frac{C_{gh}}{m_c}
\;+\; \beta^2\,\Sigma_h \Bigl(\frac{1}{m_c} + \frac{1}{m_p}\Bigr),
\label{eq:vbeta}
\end{equation}
which is minimized by
\begin{equation}
\beta^\ast \;=\; \frac{C_{gh}}{\Sigma_h}\cdot\frac{m_p}{m_p + m_c},
\qquad\text{giving}\qquad
V(\beta^\ast) \;=\; \frac{\Sigma_g}{m_c}
\Bigl(1 - \rho^2\,\frac{m_p}{m_p + m_c}\Bigr).
\label{eq:betastar}
\end{equation}
\end{lemma}

\begin{proof}
Three facts about batch means set up the computation. (1)~Averaging $m$ i.i.d.\ examples divides total variance by $m$: in $\mathbb{E}\lVert \bar g_C - \mu \rVert^2$, deviations of different examples are independent, so their inner products vanish in expectation and only the $m$ diagonal terms survive. Hence $\bar g_C$, $\bar h_C$, $\bar h_P$ have variances $\Sigma_g/m_c$, $\Sigma_h/m_c$, $\Sigma_h/m_p$. (2)~By the same diagonal argument,
$\bar g_C$ and $\bar h_C$ i.e. means over the \emph{same} examples, have covariance
$\mathbb{E}\langle \bar g_C - \mathbb{E}\bar g_C,\, \bar h_C - \mathbb{E}\bar h_C\rangle = C_{gh}/m_c$. (3)~$\bar h_P$ is computed on examples independent of $C$, so it has zero covariance with both.

Now expand $V(\beta)$ bilinearly around the means. Since
$\mathbb{E}\hat\mu_\beta = \mu$, the noise $V(\beta)$ is the squared deviation of $\hat\mu_\beta$ from its own mean; that deviation is a sum of three vectors, one per batch mean:
\begin{align}
V(\beta)
&= \mathbb{E}\,\bigl\lVert
   (\bar g_C - \mathbb{E}\bar g_C)
   + \beta\,(\bar h_P - \mathbb{E}\bar h_P)
   - \beta\,(\bar h_C - \mathbb{E}\bar h_C)
   \bigr\rVert^2
\nonumber\\[3pt]
&= \mathrm{Var}(\bar g_C)
 + \beta^2\,\mathrm{Var}(\bar h_P)
 + \beta^2\,\mathrm{Var}(\bar h_C)
 + 2\beta\,\mathrm{Cov}(\bar g_C, \bar h_P)
 - 2\beta\,\mathrm{Cov}(\bar g_C, \bar h_C)
 - 2\beta^2\,\mathrm{Cov}(\bar h_P, \bar h_C),
\label{eq:vexpand}
\end{align}
where $\mathrm{Var}(u) = \mathbb{E}\lVert u - \mathbb{E}u\rVert^2$
and $\mathrm{Cov}(u, v) = \mathbb{E}\langle u - \mathbb{E}u,\,
v - \mathbb{E}v\rangle$. Every term now matches one of the facts above. Fact~(1) supplies the three variances; fact~(2) supplies $\mathrm{Cov}(\bar g_C, \bar h_C) = C_{gh}/m_c$; fact~(3) sets both covariances involving $\bar h_P$ to zero. Substituting these values into Equation~\ref{eq:vexpand} and collecting powers of $\beta$ gives Equation~\ref{eq:vbeta}.

Finally, Equation~\ref{eq:vbeta} is an upward-opening parabola in $\beta$, so its minimum is where the derivative vanishes:
\[
-2\,\frac{C_{gh}}{m_c}
+ 2\beta\,\Sigma_h\Bigl(\frac{1}{m_c} + \frac{1}{m_p}\Bigr) = 0
\qquad\Longleftrightarrow\qquad
\beta = \frac{C_{gh}/m_c}{\Sigma_h\,(1/m_c + 1/m_p)}
= \frac{C_{gh}}{\Sigma_h}\cdot\frac{m_p}{m_p + m_c},
\]
which is $\beta^\ast$. Substituting it back into
Equation~\ref{eq:vbeta} and writing
$C_{gh}^2 = \rho^2\,\Sigma_g \Sigma_h$ yields $V(\beta^\ast)$.
\end{proof}

\paragraph{Reading the result.} The optimal coefficient is a product of two meaningful factors. The first, $C_{gh}/\Sigma_h$, is the slope of the best linear fit of $g$ on $h$: exactly how least-squares
regression would rescale the prediction before trusting it. The second, $m_p/(m_p + m_c)$, discounts the correction for being estimated from finitely many examples. The limits behave as they should. A useless predictor ($\rho \to 0$) drives $\beta^\ast \to 0$ and the estimator falls back to plain $\bar g_C$: the fleet can never hurt. A predictor that is faithful \emph{and} correctly scaled ($\rho \to 1$ with $C_{gh}/\Sigma_h \to 1$), on a large prediction batch, drives $\beta^\ast \to 1$; correlation alone does not, since a predictor that tracks $g$ perfectly at twice its scale has $\rho = 1$ but $\beta^\ast \approx 1/2$. The prize is read off Equation~\ref{eq:betastar}: for $m_p \gg m_c$ the noise shrinks to $(1-\rho^2)\,\Sigma_g/m_c$, i.e.\ the fleet makes the exact batch act $1/(1-\rho^2)$ times larger---the equivalent-batch-size currency in which we price everything \citep{large-batch}. At our deployed batch sizes ($m_c = 16$, $m_p = 332$) and fidelity at synchronized weights ($\hat\rho^2 \approx 0.98$), Equation~\ref{eq:betastar} gives about a $16\times$ multiplier on the control batch. Averaged over a broadcast period it is lower, and by how much depends on the cadence: $15\times$ in the fine-tuning window, where fidelity is flat, but about $7\times$ in the fresh window, where $\hat\rho^2$ falls to $0.8$ by the eighth step (Table~\ref{tab:staleness}): roughly $4\times$ and $2\times$ the nominal batch of 64 respectively. 
Because our predictor executes the exact backward in reduced precision rather than learning a surrogate, its scale tracks $g$ by construction, so we also run a fixed $\beta = 1$ variant as the headline configuration; it is equally unbiased and needs no statistics at all.\footnote{Note that this is different from treating the predicted gradients same as true gradients.}

\paragraph{Estimating $\beta^\ast$ without breaking unbiasedness.} The moments in equation~\ref{eq:moments} must themselves be estimated, and here one subtlety matters. The unbiasedness argument of Section~\ref{sec:prelim} treats $\beta$ as a constant; if $\beta$ is instead computed from the \emph{current} batch, it becomes correlated with the very correction it multiplies, and the product can acquire bias. The fix is to let $\beta$ depend only on the \emph{past}: conditioned on everything up to the previous step, $\beta$ is again a constant and unbiasedness holds at every step. Concretely, each step (1)~reads $\beta$ off exponential moving averages (decay $0.98$) of $C_{gh}$, $\Sigma_h$, $\Sigma_g$;
(2)~forms $\hat\mu_\beta$ and takes the optimizer step; and only then (3)~updates the averages with the current control batch's empirical moments. These were centered products of the paired $(g_i, h_i)$ with the usual
$1/(m_c{-}1)$ normalization, computed entirely in factored form via Equation~\ref{eq:gram}, so no per-example gradient is ever materialized. These are the per-example covariance statistics that the cheap forward of Section~\ref{sec:cheap-forward} captures on the control batch.

Two implementation details complete the recipe. We keep one $\beta$ \emph{per weight matrix} rather than one global scalar: gradient magnitudes differ by orders of magnitude across layers, so a single pooled coefficient would be dominated by the largest blocks, while per-block moments cost nothing extra in factored form. And we clip $\beta$ to $[0, 2]$: the clip is inactive in normal operation but bounds the damage while the moving averages warm up; thereafter the statistics defend themselves in the sense that a block whose predictions degenerate into noise sees its $C_{gh}$, and with it its $\beta$, decay to zero, automatically muting the fleet for that block.

\section{Related Work}
\label{sec:related}

\paragraph{Gradients without backward.}
Zeroth-order stochastic approximation estimates gradients from two loss evaluations alone
\citep{spall1992multivariate}, and MeZO adapts this in a way that achieves a smaller memory footprint \citep{mezo} while \citet{baydin2022gradients} enhances the method with exact forward mode derivatives. \citet{renscaling} perturb activations rather than weights. In both families the gradient estimate is unbiased but has high variance, which averaging over samples can reduce. Our predictor has the opposite error profile: it executes the exact reverse-mode computation using inference-compatible operations, so its error is systematic, arising from reduced precision and weight staleness, and does not decrease with averaging. The estimator of Section~\ref{sec:cv} is designed to remove this systematic error in expectation.

\paragraph{Approximate backward passes.}
A separate line of work replaces the exact backward pass with an approximation that is consumed directly. Synthetic gradients train auxiliary models to predict a module's error signal, allowing layers to update without waiting for the full backward pass \citep{jaderberg2017decoupled,czarnecki2017understanding}; true gradients are used only as occasional regression targets for the predictor, and the resulting update is biased whenever the prediction is imperfect. Feedback alignment replaces the transposed weight matrices with fixed random matrices \citep{lillicrap2016random,nokland2016direct}, and low-precision training executes the backward pass in reduced-precision arithmetic \citep{banner2018scalable,micikevicius2022fp8}. In all of these
methods the approximation error enters the update as bias. Our predictor combines elements of both the synthetic gradient approaches and the low-precision methods. Like synthetic gradients, it acts as an
inexpensive stand-in for the true gradient, computed separately from the trainer and therefore at possibly stale weights; like low-precision training, it obtains this stand-in by executing the exact reverse-mode computation in reduced-precision arithmetic, rather than by learning a parametric surrogate. The difference from both lines is in how the output gradient is used. We do not apply predicted gradients directly; the estimator of Section~\ref{sec:cv} uses a small number of exact gradients to remove the prediction's systematic error in expectation, and the ablation in Section~\ref{sec:frontier} quantifies the cost of direct consumption.

\paragraph{Control variates and variance reduction.}
Control variates are a standard tool for reducing the variance of gradient estimates,  for example in policy-gradient methods \citep{greensmith2004variance}. In finite-sum optimization, SVRG and SAGA subtract a correlated reference gradient and add back its exactly known mean  \citep{NIPS2013_ac1dd209,defazio2014saga}. Both estimators are unbiased, and their effectiveness depends on the correlation between the reference and the current gradient.  \citet{defazio2019ineffectiveness} show that in deep networks this correlation decays quickly as the iterates move away from the snapshot, so the variance reduction is small relative to the cost of maintaining the reference. Our estimator uses the same principle with a different reference: a quantized reverse-mode prediction that is recomputed at near-current weights at every step, evaluated on both the control batch and a larger independent batch, and billed at the fleet rate. Because the reference is inexpensive, it can be kept close to the current iterate, and the resulting correlation is
high on our testbed ($\hat\rho^2 \approx 0.98$); Section~\ref{sec:horizon} measures how it degrades with weight staleness. A further difference is that the reference mean is not known exactly, as in SVRG, but estimated on the prediction batch, whose size enters the variance expression of Lemma~\ref{lem:beta}.

\paragraph{Distributed training and gradient compression.}
Federated, local-SGD, and low-communication methods reduce synchronization by performing local updates \citep{mcmahan2017communication,stichlocal,diloco}; SWARM and Bamboo extend training to unreliable or preemptible workers \citep{ryabinin2023swarm,thorpe2023bamboo}. In all of these methods, each worker executes an exact backward pass on training-capable hardware. In our setting the fleet does not: it returns predicted batch-gradient means and control statistics, while exact gradients and optimizer state remain on the trainer. Our
use of stale weights also differs from asynchronous SGD \citep{recht2011hogwild,dean2012large}, where stale gradients are applied directly. Here a stale prediction is not consumed directly: the same predictor appears in both correction terms, so staleness affects the correlation and hence the variance, but not the expectation of the gradient estimate. Gradient compression \citep{vogels2019powersgd} encodes an already-computed gradient and is
complementary to our protocol.

\paragraph{Data selection.}
Importance sampling, proxy selection, and learnability-based prioritization use
inexpensive forward passes to decide which examples receive an exact update
\citep{katharopoulos2018not,colemanselection,mindermann2022prioritized}. We instead use the same compute to obtain approximate gradients for additional examples. Since the two approaches draw on the same budget, we include a matched-budget selection baseline; in our experiments, gradient prediction outperforms it at nearly all fleet prices we consider (Section~\ref{sec:frontier}).

\paragraph{Effective batch size.}
Studies of large-batch training relate gradient noise to the useful batch size and find that this relationship depends on the workload and the optimizer
\citep{large-batch,shallue2019measuring,zhang2019algorithmic}. We use equivalent batch size as the common unit of comparison, but obtain the variance reduction from predicted gradients, account for their cost explicitly in the cost model of
Section~\ref{sec:prelim}, and distinguish the statistical variance reduction from the portion realized as training progress by the optimizer  (Section~\ref{sec:absorption}).

\section{Experimental Results}\label{sec:results}

We put four questions to the method. 
\begin{enumerate}
    \item How does simulated ledger cost to a target loss vary with the fleet price $\gamma$ (Section~\ref{sec:frontier})?
    \item Why does the choice of optimizer decide between winning and losing (Section~\ref{sec:absorption})?
    \item How long does the win last (Section~\ref{sec:horizon})?
    \item And does it survive a change of scale, dataset, and model family (Section~\ref{sec:transfer})?
\end{enumerate}

\begin{figure}[t]
\centering
\includegraphics[width=.8\textwidth]{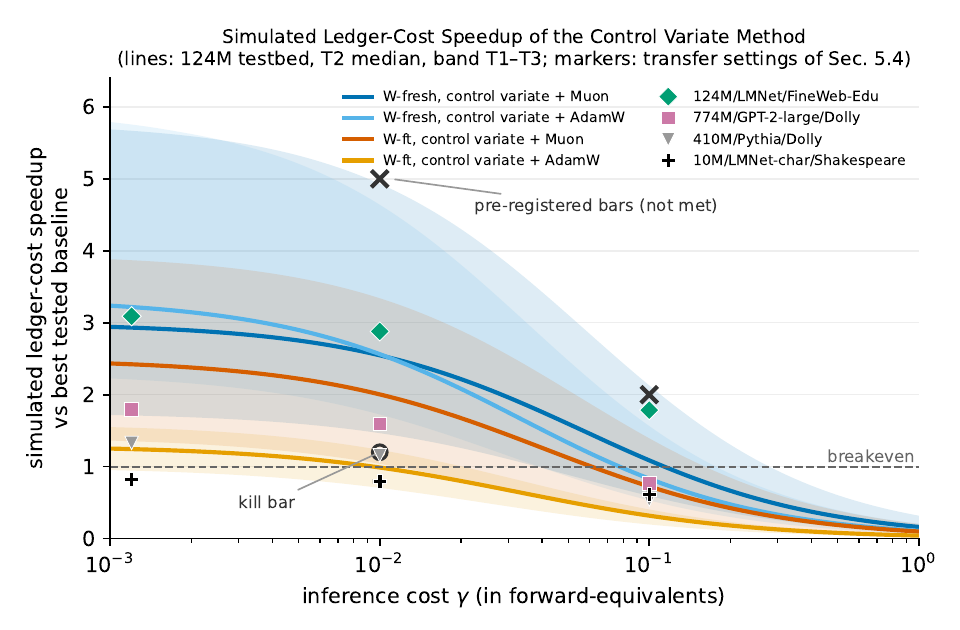} 
\caption{Simulated ledger-cost ratio (median cost of the lowest-cost tested baseline divided by the arm's median cost) to target $T_2$ as a function of billing rate $\gamma$ (band: $T_1$-$T_3$); values above one favor the method. Transfer settings from Section~\ref{sec:transfer} are overlaid as points.
\label{fig:frontier}}
\end{figure}

\paragraph{Setup.} Our main testbed is a 124M-parameter GPT-2-class model trained on TinyStories. We evaluate in two windows chosen to be as different as possible: \emph{W-fresh} continues mid-training on fresh data, with weights broadcast to the fleet every 8 steps, and \emph{W-ft} fine-tunes a late checkpoint on held-out data at a low
learning rate, broadcasting every 32 steps. Each training configuration (an \emph{arm}) gets the same budget of 120 scarce-GPU seconds and its own tuned learning rate, and we record what it costs to reach each of three validation-loss targets $T_1, T_2, T_3$ (set at fixed fractions of the best tested baseline's descent). The cost of a run at billing rate $\gamma$ is
\begin{equation}
c_\gamma \;=\; t_{\text{scarce}} \;+\; \gamma \cdot F \cdot
t_{\text{fwd}},
\label{eq:cost}
\end{equation}
where $t_{\text{scarce}}$ is the scarce-GPU time spent at the point the target is reached, $F$ counts the fleet's forward-equivalents up to that point, and $t_{\text{fwd}}$ is the duration of one forward on the scarce GPU. \emph{Speedup} is the best tested baseline's median cost divided by the arm's median cost (3--5 seeds). The baselines are tuned AdamW, Muon, small-batch AdamW, and a selector that spends the same fleet budget on choosing data rather than predicting gradients.\footnote{Use forward to compute loss, sample proportionally to loss, train with a weight inverse to the
probability of sampling.} We did not run a small-batch Muon baseline or sweep batch size for every optimizer, so ``best baseline'' means the best of these four at their tuned learning rates; the batch-size axis is examined separately, for both AdamW and Muon, in Section~\ref{sec:absorption}.
Learning-rate tuning, the correctness gates that every experiment must pass first, and all reproduction details are in Appendix~\ref{app:exp-details}.

\paragraph{The predictor in two numbers.} Everything below is priced by the predictor's fidelity and its cost. Fidelity: on this testbed the int8 prediction explains $\hat\rho^2 = 0.982$-$0.988$ of per-example gradient variance. Cost: the fleet meters its own work against a fixed forward unit and bills $c_h = 7.4$ forward-equivalents per example served; this metered rate generates every $\gamma$ column in the paper. The predictor kernel alone measures $7.0$ offline, and a conservative chain without fused attention kernels measures $9.2$; Appendix~\ref{app:robustness} re-bills every run at the latter. At $\gamma = 0.01$, this amounts to about 3\% of one training step. A full cost breakdown is in Appendix~\ref{app:exp-details}.

\subsection{The frontier}\label{sec:frontier}
Figure~\ref{fig:frontier} is the central result\footnote{No curve fitting is involved in Figure~\ref{fig:frontier}. Because the ledger cost of Equation~\ref{eq:cost} is a linear function of $\gamma$ with fixed, measured meters, the same runs can be re-billed exactly at any price: each line evaluates the median cost of the lowest-cost tested baseline divided by the arm's median cost at target $T_2$ on a dense grid of $\gamma$ values, and its smoothness is a property of the formula, not of any smoothing.}: the simulated ledger-cost ratio as a function of the fleet billing rate. Table~\ref{tab:frontier} gives the exact ratios at the reported operating points. The advantage is largest in the idealized free-fleet limit, remains above parity for some mid-training settings at $\gamma = 0.01$, and largely disappears by $\gamma = 0.1$. The inferred parity crossing varies substantially across targets. These are compute-ledger comparisons against the tested baseline set, not end-to-end wall-clock speedups.

\begin{table}[t]
\centering\small
\begin{tabular}{llcccc}
\toprule
window & arm & seeds & free ($\gamma{=}0$) & $\gamma{=}0.01$ &
$\gamma{=}0.1$ \\
\midrule
W-fresh & control variate (AdamW) & 3 & 2.3,\,3.3,\,6.0 & 1.7,\,2.6,\,4.6 & 0.5,\,0.8,\,1.6 \\
W-fresh & control variate (Muon) & 5 & 1.8,\,3.0,\,5.8 & 1.5,\,2.5,\,4.9 & 0.6,\,1.1,\,2.1 \\
W-ft & control variate (AdamW) & 3 & 1.0,\,1.3,\,1.6 & 0.7,\,1.0,\,1.2 & 0.2,\,0.3,\,0.4 \\
W-ft & control variate (Muon) & 5 & 1.4,\,2.5,\,4.0 & 1.0,\,2.0,\,3.3 & 0.3,\,0.7,\,1.4 \\
\bottomrule
\end{tabular}
\caption{Simulated ledger-cost speedup vs the best tested baseline
(of four tuned baselines) at targets $[T_1, T_2, T_3]$ (medians over
seeds).\label{tab:frontier}}
\end{table}

Before running anything we committed two decision rules: an ambitious success bar ($\ge$2$\times$ at $\gamma{=}0.1$ \emph{and} $\ge$5$\times$ at $\gamma{=}0.01$) and a kill bar ($\le$1.2$\times$ at $\gamma{=}0.01$). The measured result lands between them, and narrowly: at its tightest target the best arm reaches 2.1$\times$ at $\gamma{=}0.1$ and 4.9$\times$ at $\gamma{=}0.01$, missing the $5\times$ leg by a hair; at the pre-registered target $T_2$ it clears neither leg, so the success bar goes unmet. The kill fired for exactly one configuration: the AdamW-routed anchor in the fine-tuning window, which clears 1.2$\times$ at only one of the three targets while the Muon-routed arm in the same window clears two. Understanding that one failure is the subject of the next subsection.

Three ablations complete the picture, all in the fine-tuning window. Removing the anchor and consuming raw predictions costs 0.2-0.8$\times$, and the arm never reaches the tightest target at all: bias, unlike variance, does not average out. The adaptive-$\beta$ estimator neither helps nor hurts much: it runs at 0.7-0.9$\times$ of the fixed $\beta = 1$ anchor, which is what Lemma~\ref{lem:beta} predicts at this fidelity.\footnote{Equation~\ref{eq:betastar} pins $\beta^\ast$ only up to the regression slope $C_{gh}/\Sigma_h$, which we did not measure, so we do not claim $\beta = 1$ is optimal and instead report both arms. Adaptive $\beta$ earns its keep not here but as protection: it is what mutes a block whose predictions degenerate. Both ablations ran one seed, against three and five for the anchored arms.} And the data-selection baseline is beaten by gradient prediction everywhere except at the loosest target at $\gamma = 0.1$: given the same fleet budget, gradients are generally worth more than data selection.

\subsection{Why the optimizer decides the verdict}
\label{sec:absorption}
Section~\ref{sec:cv} showed that the estimator behaves like training with a batch $\zeta$ times larger. But a larger effective batch only helps if the optimizer turns lower gradient noise into faster progress, and optimizers differ in how well they do that. We measure the conversion directly: run anchored training with a variance reduction nominally worth $\zeta \in \{2, 4\}$, find the exact batch size that reaches the targets in the same number of steps, and call the shortfall $D$ (nominal $\zeta$ over realized $\zeta$, so $D = 1$ means nothing is wasted)~\citep{large-batch}.

Table~\ref{tab:absorption} shows the result, and it depends on the phase. While the loss is still descending, AdamW wastes 16-67\% of the win, whereas Muon gains \emph{more} than the variance accounting promised ($D = 0.77$, $0.58$); in the shallow fine-tuning window the ordering reverses. This is why the headline arm routes the estimate through Muon~\citep{jordan2024muon}; the reversal recurs in the transfer settings of Section~\ref{sec:transfer}, where we flag it as an open question. (SGD-family arms convert almost none of the win: realized $\zeta \le 1$ against a nominal 2 or 4 and are unstable at production learning rates on this testbed.)

\begin{table}[t]
\centering\small
\begin{tabular}{lcccc}
\toprule
 & \multicolumn{2}{c}{AdamW} & \multicolumn{2}{c}{Muon} \\
phase & $\zeta{=}2$ & $\zeta{=}4$ & $\zeta{=}2$ & $\zeta{=}4$ \\
\midrule
mid-training & 1.16 & 1.67 & 0.77 & 0.58 \\
fine-tune (shallow) & 1.02 & 0.69 & 1.10 & 1.56 \\
\bottomrule
\end{tabular}
\caption{Absorption $D$ = nominal $\zeta$ / realized $\zeta$
(1.0 = the optimizer converts the full variance win; larger is
worse).\label{tab:absorption}}
\end{table}

\subsection{How long the win lasts}\label{sec:horizon}

\begin{figure}[t]
\centering
\includegraphics[width=.78\textwidth]{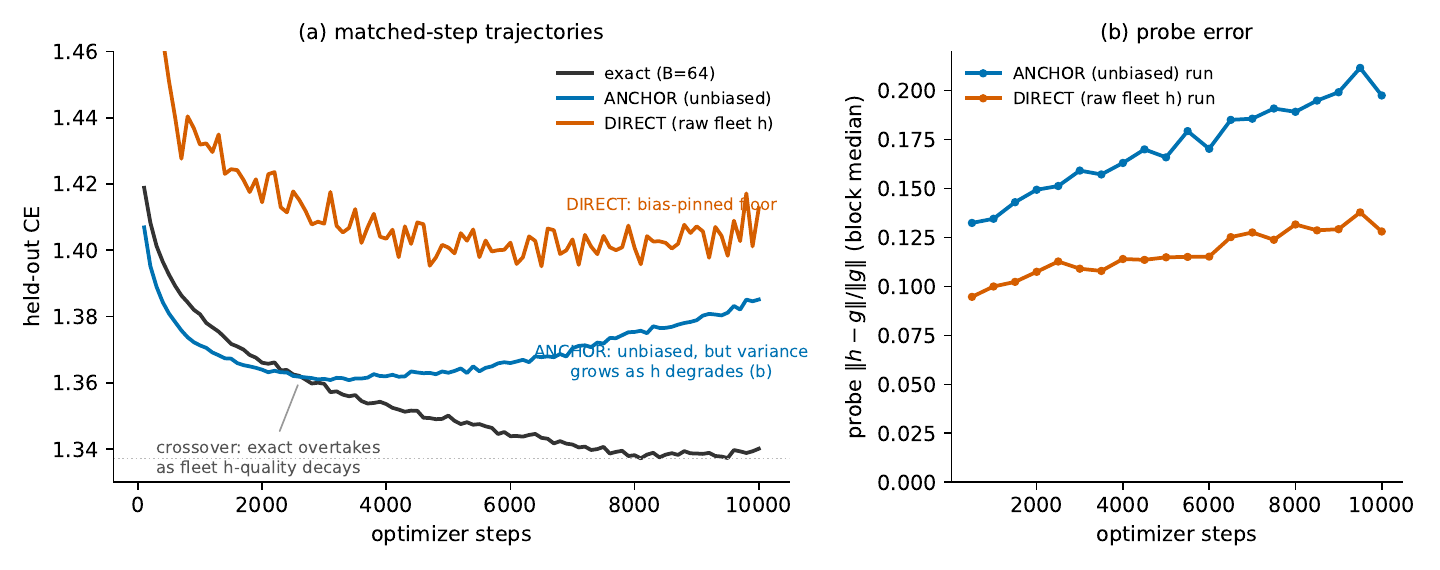}
\caption{Matched 10k-step runs (AdamW, fixed LR): exact $B{=}64$,
anchored, and raw prediction consumption; right panel: the
predictor's probe error along the trajectory.\label{fig:drift}}
\end{figure}

The training windows are short by design. Figure~\ref{fig:drift} runs the estimator for 10{,}000 steps at a fixed learning rate, against matched exact training and against raw prediction consumption. Early, the anchored run beats exact training step for step: the variance win is real. It bottoms out near step 3{,}400 and then drifts up, as the predictions degrade away from the last broadcast weights (probe error\footnote{The \emph{probe error} is a diagnostic measured every 500 steps, outside the training update. The fleet's weight copy is first synced to the current weights (so weight staleness contributes nothing), then the prediction $h$ and the exact gradient $g$ are computed on the same 64 fresh examples; the reported number is the relative error $\lVert h - g\rVert/\lVert g\rVert$ per weight matrix, median across matrices.} $0.13 \to 0.20$). The anchor turns that decay into variance rather than bias, so the edge fades gracefully; however, at a constant learning rate variance sets the loss floor, not just the speed, so the curve rises with it. Raw consumption, by contrast, sits $0.04$-$0.10$ cross-entropy units above exact training throughout; no amount of training rescues a biased update. The deployment recipe follows: take the initial win, watch the online $\hat\rho^2$, and hand back to exact training when it decays. Panel (b) should not be read as a contest between the two: the probe measures how hard prediction is wherever a run has got to, and the anchored run, being further down the loss curve, where gradients carry less signal, is simply probing harder ground.

A second axis of degradation is weight age. Table~\ref{tab:staleness} gives the prediction fidelity when the fleet's weights are $\Delta$ steps old, with the trajectory held fixed so that staleness is the only variable. At fine-tuning learning rates it is a non-issue: fidelity is flat at $0.98$ out to $\Delta = 16$, while at production rates it falls to $0.79$ by $\Delta = 8$. This is what sets the broadcast cadence: 32 steps in the fine-tuning window, 8 in the fresh one.

\begin{table}[t]
\centering\small
\begin{tabular}{lccccccc}
\toprule
$\Delta$ (steps stale) & 0 & 1 & 2 & 4 & 8 & 16 & 100 \\
\midrule
W-ft (LR $10^{-5}$) & .982 & .982 & .982 & .982 & .982 & .980 & .969 \\
W-fresh (production LR) & .988 & .976 & .945 & .883 & .794 & .704 & .484 \\
\bottomrule
\end{tabular}
\caption{Prediction fidelity $\hat\rho^2$ vs weight staleness
$\Delta$.\label{tab:staleness}}
\end{table}

\subsection{Where it transfers}\label{sec:transfer}

\begin{figure}[t]
\centering
\includegraphics[width=\textwidth]{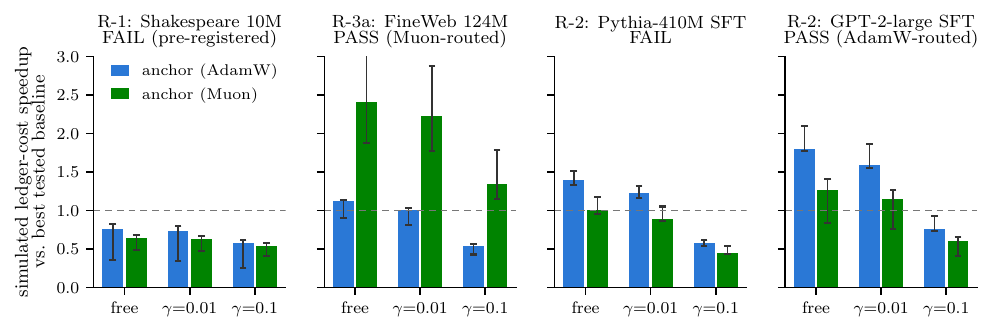}
\caption{Transfer verdicts (bars: median simulated ledger-cost ratio at $T_2$; whiskers: range over targets; dashed line: parity). Note the routing reversal: Muon wins the pretraining-style window (FineWeb) and loses both SFT windows.\label{fig:rphase}}
\end{figure}

Finally, we re-ran the full protocol in four settings chosen to probe its boundaries, holding the estimator fixed ($m_c{=}16$, sync-8, $\beta{=}1$) and sizing only the prediction batch to each model: a 10M character-level model trained from scratch on Shakespeare; the same 124M architecture we used before pretrained and windowed on FineWeb-Edu; and supervised fine-tuning of two public checkpoints, Pythia-410M and GPT-2-large (774M), on Dolly-15k. Each setting had a pre-committed transfer bar ($\ge$1.5$\times$ free-limit and $\ge$1.2$\times$ at $\gamma{=}0.01$, on at least 2 of 3 targets). Figure~\ref{fig:rphase} and Table~\ref{tab:rphase} give the verdicts: two pass, two fail. The failures mark the boundaries of the method.

\begin{table}[t]
\centering\small
\begin{tabular}{llllll}
\toprule
setting & scale & fleet $\hat\rho^2$ & bar & $\gamma{=}0$ & $\gamma{=}0.01$ \\
\midrule
Shakespeare (char, scratch) & 10M & 0.553--0.634 & FAIL & 0.4--0.8$\times$ & 0.3--0.8$\times$ \\
FineWeb-Edu (fresh window) & 124M & 0.959--0.979 & \textbf{PASS} (Muon) & 1.9--3.1$\times$ & 1.8--2.9$\times$ \\
Pythia-410M (Dolly SFT) & 410M & 0.9774$^\dag$ & FAIL & 1.3--1.5$\times$ & 1.2--1.3$\times$ \\
GPT-2-large (Dolly SFT) & 774M & \textbf{0.9997}$^\dag$ & \textbf{PASS} (AdamW) & 1.8--2.1$\times$ & 1.6--1.9$\times$ \\
\bottomrule
\end{tabular}
\caption{The transfer map (winning control-variate arm per setting;
ranges over targets). The Pythia/GPT-2 pair is the controlled contrast:
similar scale, same data, opposite verdicts, tracking fleet fidelity
exactly. $^\dag$fidelity carried from the offline probe; not re-measured
in this run.\label{tab:rphase}}
\end{table}

Together the four verdicts say when the method applies. (i)~\emph{The exact backward must be expensive enough.} At 10M a baseline step takes about 12ms, and the anchored step costs three times that. This amounts to too little work to amortize the estimator's fixed per-step machinery, so nothing can win. (Interestingly, the anchored arm still reaches the lowest median final loss, statistically tied with the best tested baseline; it loses the race, not the endpoint.) (ii)~\emph{Prediction fidelity is a property of the model family, not of scale.} At comparable size and on the same task, GPT-2-large's predictions are near-perfect (0.9997, pass), while the outlier structure of the Pythia residual stream, the same structure that makes int8 unusable on that family~\citep{inteight}, caps fidelity at 0.98: enough to train, not enough to win. (iii)~\emph{There must be headroom to spend on the ledger.} The predictor costs roughly a dozen forward-equivalents per example at these scales, only somewhat more than on our main testbed; what changes is how much ledger-cost headroom there is to bill against. The transfer settings start from a much smaller free-limit win, and a $\gamma{=}0.1$ ledger costs more than that win is worth. (iv)~\emph{The optimizer routing must match the regime, and we cannot yet predict how.} The winning route flips from setting to setting, and even between the two windows of our own testbed, with no pattern we can defend, a reversal visible across Table~\ref{tab:frontier} and Figure~\ref{fig:rphase} that we flag as an open question rather than explain.

The verdict is also robust to how we measured our own constants, though not equally in both directions: re-billing the runs under adverse kernel prices and favorable overhead leaves the win at $\gamma{=}0.01$ intact in every case, never dropping below roughly $1.9\times$. The loss at $\gamma{=}0.1$ is the fragile half. In the fresh window it sits close to break-even and tips to either side depending on the assumed price, which is the same marginality Figure~\ref{fig:frontier} shows (Appendix~\ref{app:exp-details}, Table~\ref{tab:sensitivity}).

\section{Conclusions}
We introduced a GPU implementation of a reduced-precision, inference-style reverse-mode predictor and a control-variate estimator that remains unbiased for any predictor under the stated sampling and coefficient assumptions. Under a simulated compute ledger, some configurations lower cost relative to the tested, non-exhaustive baselines in short training windows when fleet work is sufficiently cheap; transfer results are mixed. The benefit is conditional on predictor fidelity, amortization, and optimizer choice. We did not test inference-only hardware, end-to-end distributed wall-clock time, communication, memory on real inference silicon, or a full optimizer-by-batch-size baseline sweep. Thus the validated result is an estimator plus a GPU-measured compute-ledger tradeoff, not demonstrated deployment on inference-only hardware or an end-to-end training speedup.

\bibliography{main}
\bibliographystyle{tmlr}

\appendix

\section{Experimental details and sanity checks}\label{app:exp-details}

\subsection{Protocol and reproduction details}\label{app:protocol}

\paragraph{Windows.} W-fresh continues the 124M testbed from its step-6k checkpoint on a fresh, disjoint region of the training corpus, with weights broadcast to the fleet every 8 steps; W-ft fine-tunes the step-20k checkpoint on held-out data at a low learning rate, with a broadcast every 32 steps. Validation loss is measured on a region disjoint from both training pools. Control-variate arms use $m_c = 16$ control examples and $m_p = 332$ prediction examples per step, at a nominal batch of 64.

\paragraph{Learning rates.} Every arm in every window is tuned on a geometric grid using 60-second pilot runs; the winner is the lowest held-out loss, ties resolved to the lower rate, and grid-edge winners extended until the winner is interior. The chosen rates, together with the success and kill bars and the per-arm predictions, were committed to \emph{before} any timed run.

\paragraph{Measurement.} The targets $T_1, T_2, T_3$ are the loss values at 25/50/75\% of the best tested baseline's own descent within that window, so they are set by the strongest tested competitor rather than by an absolute number carried in from elsewhere. Every run logs its own scarce seconds and fleet forward-equivalents at each log tick. This means that any run can be re-billed at any $\gamma$ after the fact; all $\gamma$ columns in the paper come from one set of runs. Timed runs get a 120-second scarce budget. Ledger-cost ratios use medians over 3--5 seeds per arm; the artifact carries full per-seed ranges, and at this $n$ we report ranges rather than parametric error bars. During fleet work the scarce clock is paused and the work is metered in forward-equivalents; communication and distributed-system latency are not included. The fleet is a service on the same box as the trainer, so the reported ratios are not end-to-end wall-clock speedups.

\subsection{Correctness gates (aka unit tests)}\label{app:gates}

Before any timed run, three kinds of checks must pass (Table~\ref{tab:gates}). The fp64 tie-back certifies that the cheap forward computes the right formulas: run in double precision with quantization off, it must reproduce autograd's per-example gradients to relative error $10^{-12}$, which it does with nearly three orders of magnitude to spare. The hostile-predictor gate certifies the estimator: we corrupt the predictor with stale weights, asymmetric rescaling, fixed bias, and int8-style rounding, and test the estimate's mean over 400 redraws. The anchored estimator stays unbiased\footnote{The $|t|$ values are standard $t$-statistics: the estimate's mean minus the known pool mean, divided by its standard
error, taken over 400 independent redraws and 6 random projection directions (worst reported).}, while consuming the raw predictions fails the same test at $|t| = 67$. The variance formula of Lemma~\ref{lem:beta} holds to better than 1\% on the same hostile
predictor, and the per-block $\beta$ correctly mutes a pure-noise block while leaving a good block near 1. Beyond these formula-level gates, every deployment must pass an end-to-end check that catches indexing and orientation bugs.

\begin{table}[t]
\centering\small
\begin{tabular}{llll}
\toprule
check & quantity & measured & bar \\
\midrule
formulas & fp64 tie-back vs autograd (124M) & $1.5\times10^{-15}$ &
$\le 10^{-12}$ \\
 & fp64 tie-back, HF families & $8.5\times10^{-14}$ & $\le 10^{-12}$ \\
unbiasedness (G1) & $|t|$, anchored / adaptive-$\beta$ &
1.87 / 1.21 & $< 4$ \\
 & $|t|$, no anchor (must fail) & 67.3 & $> 6$ \\
variance law (G2) & rel.\ deviation from Eq.~\ref{eq:vbeta} &
0.005 & $< 0.35$ \\
veto (G3) & $\beta$ on noise / good block &
0.001 / 0.893 &
$<0.15$ / 0.6--1.3 \\
\bottomrule
\end{tabular}
\caption{Correctness gates, run before every timed experiment.
\label{tab:gates}}
\end{table}

\subsection{The predictor's price in detail}\label{app:cost}

\begin{figure}[t]
\centering
\includegraphics[width=.55\textwidth]{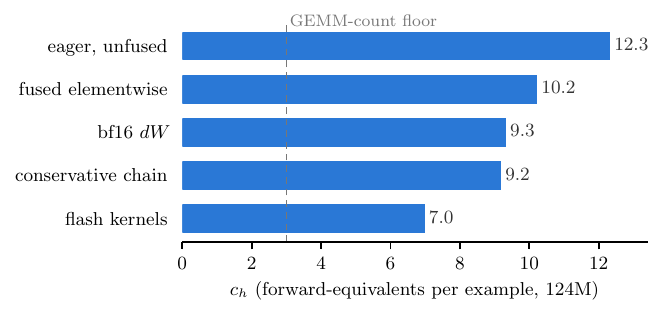}
\caption{The predictor's price $c_h$ at 124M, per optimization stage;
the dashed line is the GEMM-count lower bound. 
\label{fig:chladder}}
\end{figure}

Figure~\ref{fig:chladder} shows $c_h$, the predictor's price in forward passes per predicted gradient, at each stage of its optimization: from a direct transcription (12.3) down through fused elementwise work (10.2) to the conservative bf16 chain, which this run measures at 9.2, and finally to a variant that calls fused attention kernels directly (7.0); counting GEMMs bounds any implementation below at about 3. The frontier runs are billed at neither ladder rung directly but at the fleet service's own live meter, $7.4$ forward-equivalents per served example, which exceeds the kernel-only $7.0$ by the service's per-job overhead and statistics slice. The price is flat in batch size (8.4 / 6.9 / 6.8 forwards at batch 16 / 64 / 332) and the two attention paths give the same predictions to three decimals, so the kernels change the price and not the fidelity. The trainer's own
per-step machinery costs $\kappa \le 1.10$ of a plain step.\footnote{$\kappa$ compares an anchored step to a plain one, counting only the work that stays on the scarce GPU: taking the exact gradients of the few control examples apart example by example, averaging them, and keeping the running statistics that set $\beta$. The first two are essentially free — the capture chain is even a little cheaper than an ordinary step, because it never builds an autograd graph — and the statistics, refreshed every few steps rather than every step, account for the remaining tenth.}

Fidelity is uniform across the network: pooled $\hat\rho^2$ is 0.988 at the step-6k checkpoint and 0.982 at step-20k, and no individual weight matrix falls below 0.94 at either. Two break-even numbers should not be confused. The idealized cost model of Section~\ref{sec:prelim} breaks even at $\gamma \approx 0.32$ with fused kernels, or $0.24$ without, whereas the measured frontier crosses parity between $\gamma \approx 0.01$ and $0.16$ in the fine-tuning window and between $0.04$ and $0.28$ in the mid-training one. The gap is exactly what the absorption factor $D$ and staleness take, as quantified in Sections~\ref{sec:absorption} and~\ref{sec:horizon}.

\subsection{Robustness of the verdict}\label{app:robustness}

Both measured constants carry an honest error bar, so we re-bill every  frontier run under the conservative kernel price and under a favorable overhead assumption (Table~\ref{tab:sensitivity}). The conservative case multiplies the metered fleet price by $9.3/7.0$, billing at about $9.8$ forward-equivalents per served example, if anything harsher than the $9.2$ we measure for that path. The two halves of the verdict are not equally robust. The win at $\gamma = 0.01$ survives every combination, in both windows, never dropping below $1.9\times$; it is not an artifact of how we priced our own kernels. The loss at $\gamma = 0.1$ is the fragile half: it holds throughout in W-ft, but in W-fresh the arm sits within about 10\% of parity and its sign depends on which kernel price we assume. That marginality is visible directly in Figure~\ref{fig:frontier}, where the $\gamma = 0.1$ end of the fresh
window straddles the break-even line.

\begin{table}[t]
\centering\small
\begin{tabular}{lcc}
\toprule
re-billing case & W-fresh: $0.01$ / $0.1$ & W-ft: $0.01$ / $0.1$ \\
\midrule
as metered ($c_h$ 7.4) & \textbf{2.55} / \textbf{1.09} & \textbf{2.00} / 0.72 \\
conservative kernels ($1.33\times$ fleet) & \textbf{2.43} / 0.90 & \textbf{1.88} / 0.59 \\
favorable overhead ($0.92\times$ scarce) & \textbf{2.74} / \textbf{1.12} & \textbf{2.14} / 0.74 \\
both & \textbf{2.60} / 0.92 & \textbf{2.01} / 0.60 \\
\bottomrule
\end{tabular}
\caption{The Muon-routed control-variate arm's simulated ledger-cost
speedup at $T_2$ when the frontier runs are re-billed under perturbed
constants (bold $=$ faster than the best tested baseline). The win at $\gamma{=}0.01$ survives every
combination; the loss at $\gamma{=}0.1$ does not, and in the fresh
window the sign depends on the assumed kernel price.
\label{tab:sensitivity}}
\end{table}

\section{Approach to statistical rigor}\label{app:stats}

Timed runs on scarce hardware are expensive, so our sample sizes are small: 3-5 seeds per arm per window. We would rather be explicit about what that does and does not support than dress it up in machinery it cannot carry. This appendix states where the confidence in our numbers
actually comes from.

\paragraph{We replicated the experiments, not just the seeds.}
Everything reported in this paper comes from a single end-to-end execution of our reproduction script, which we call run~1. We then executed the whole experimental protocol a second time, re-running every timed arm  at a different seed, and analyzed the result with
the same code. Run~2 shares the token packs and the pretrained checkpoints with run~1, so it is a replication of the experiments rather than of the pretraining; it is also not a fourth seed folded into a median, and we report it as a check on run~1 rather than averaging the two together. The two agree on every qualitative claim in the paper: the same optimizer wins each window, the same single configuration trips the kill bar, and all eight transfer verdicts (two arms in each of four settings) come out identically, including both passes and both failures.
The absorption factors of Section~\ref{sec:absorption} likewise match to within about a tenth in every cell. Quantitatively, the ledger-cost rations (speedups) at the central target $T_2$ agree to within 5\%, with a median disagreement of about 2\%. The looser and tighter targets $T_1$ and $T_3$ move more, up to about 19\%, which is what one should expect, since they sit where a loss curve is steep or nearly flat, and a small shift in the curve there translates into a large shift in the
time to reach it. Run~2 is slightly more favorable to our method than run~1 in about two thirds of the cells; we report run~1 throughout.

\paragraph{We report ranges, not error bars.}
At three to five seeds, a parametric confidence interval would be a statement about a normal approximation we cannot check, not about our data. We therefore report medians over seeds together with the full per-seed spread in the artifacts. Seed-to-seed variation in the frontier
runs turns out to be small relative to the effects we are claiming; for instance, the five seeds of the Muon-routed arm at $T_2$ and $\gamma = 0.01$ land within a few percent of each other in both windows. The honest summary is that seeds are not the dominant uncertainty here. The dominant uncertainties are the choice of target and the price we assign to our own kernels, and both are reported as explicit spans rather than folded into a single interval.

\paragraph{We decided in advance what would count as success.}
The statistical weight in this paper rests on decision discipline rather than on interval estimates. Before any timed run, we committed to the learning rates, to a success bar, to a kill bar, and to a predicted range for each arm. Outcomes that missed their predicted ranges are not dropped or reweighted, and the pre-registered failures in Section~\ref{sec:transfer}
are reported as failures.

\paragraph{Three things the figures are not.}
First, the shaded bands in Figure~\ref{fig:frontier} are not confidence intervals. They are the envelope over the three loss targets $T_1$-$T_3$, with seed variation already collapsed into the medians; a wide band means the ledger-cost ratio (speedup) depends on how far you want to train, not that the measurement is noisy. Second, the columns at different $\gamma$ are not independent measurements. Each run logs its own scarce seconds and fleet forward-equivalents, and every $\gamma$ column is a deterministic re-billing of those same logs, so comparisons across
$\gamma$ are exactly paired and carry no extra sampling error. Third, the targets are not absolute constants: they are set at fixed fractions of the best baseline's own descent within each run, which is what makes run~1 and run~2 comparable despite landing at slightly different
absolute losses.

\paragraph{Where we do use a statistical test, we use it properly.}
One claim in the paper is genuinely statistical: that the control variate leaves the gradient estimate unbiased. We test it with explicit $t$-statistics against pre-stated pass/fail thresholds, under a deliberately hostile predictor, together with a negative control that must fail; it does fail by a wide margin (Table~\ref{tab:gates}). These gates are deterministic given their seed and reproduced bit-for-bit across both runs, so they certify the implementation rather than sampling the training process.

\paragraph{What this still does not cover.}
Two limits are worth stating plainly. Our per-setting seed counts would not support a claim of a small effect, and we make none; the effects we report are factors, not percentages, except at the boundaries where we say so. And a handful of quantities: the upper rungs of the cost ladder, the fidelity probes for the two Hugging Face families, and the sync-period envelope are carried from our earlier runs rather than re-measured by the reproduction.

\end{document}